\documentclass[runningheads]{llncs}
\usepackage{graphicx}
\usepackage{amsmath,amssymb} 
\usepackage{color}
\usepackage[width=122mm,left=12mm,paperwidth=146mm,height=193mm,top=12mm,paperheight=217mm]{geometry}
\usepackage{times}
\usepackage{epsfig}
\usepackage{bbm}
\usepackage{float}

\usepackage{tabularx}
\usepackage{booktabs}
\usepackage{multirow}
\usepackage{xspace}
\usepackage{xcolor, colortbl}
\usepackage{caption} 
\usepackage{algorithm2e}
\usepackage{arydshln}

\newcommand{\minisection}[1]{\textbf{#1}\hspace{0.3em}}

\newcommand{\ones}{\mathbbm{1}}
\newcommand{\elprod}{\odot}

\usepackage{etoolbox}
\newcommand{\repthanks}[1]{\textsuperscript{\ref{#1}}}
\makeatletter
\patchcmd{\maketitle}
  {\def\thanks}
  {\let\repthanks\repthanksunskip\def\thanks}
  {}{}
\patchcmd{\@maketitle}
  {\def\thanks}
  {\let\repthanks\@gobble\def\thanks}
  {}{}
\newcommand\repthanksunskip[1]{\unskip{}}
\makeatother

\begin{document}
\pagestyle{headings}
\mainmatter
\def\ECCV16SubNumber{449}

\title{Clockwork Convnets for Video Semantic Segmentation}

\titlerunning{Clockwork Convnets for Video Semantic Segmentation}

\authorrunning{E. Shelhamer$^*$, K. Rakelly$^*$, J. Hoffman$^*$, and T. Darrell}

\author{Evan Shelhamer\thanks{Authors contributed equally.\protect\label{auth}}\hspace{2em}
    Kate Rakelly\repthanks{auth}\hspace{2em}
    Judy Hoffman\repthanks{auth}\hspace{2em}
    Trevor Darrell\\
    {\tt\small \{shelhamer,rakelly,jhoffman,trevor\}@cs.berkeley.edu}
}

\institute{UC Berkeley}

\maketitle

\begin{abstract}
Recent years have seen tremendous progress in still-image segmentation; however the na\"ive application of these state-of-the-art algorithms to every video frame requires considerable computation and ignores the temporal continuity inherent in video.
We propose a video recognition framework that relies on two key observations: 1) while pixels may change rapidly from frame to frame, the semantic content of a scene evolves more slowly, and 2) execution can be viewed as an aspect of architecture, yielding purpose-fit computation schedules for networks.
We define a novel family of ``clockwork'' convnets driven by fixed or adaptive clock signals that schedule the processing of different layers at different update rates according to their semantic stability.  
We design a pipeline schedule to reduce latency for real-time recognition and a fixed-rate schedule to reduce overall computation.
Finally, we extend clockwork scheduling to adaptive video processing by incorporating data-driven clocks that can be tuned on unlabeled video.
The accuracy and efficiency of clockwork convnets are evaluated on the Youtube-Objects, NYUD, and Cityscapes video datasets.

\end{abstract}

\section{Introduction}
\label{sec:intro}
Semantic segmentation is a central visual recognition task.
End-to-end convolutional network approaches have made progress on the accuracy and execution time of still-image semantic segmentation, but video semantic segmentation has received less attention.
Potential applications include UAV navigation, autonomous driving, archival footage recognition, and wearable computing.
The computational demands of video processing are a challenge to the simple application of image methods on every frame, while the temporal continuity of video offers an opportunity to reduce this computation.

Fully convolutional networks (FCNs) \cite{fcn,eigen2015predicting,fischer2015flownet} have been shown to obtain remarkable results, but the execution time of repeated per-frame processing limits application to video.
Adapting these networks to make use of the temporal continuity of video reduces inference computation while suffering minimal loss in recognition accuracy.
The temporal rate of change of features, or feature ``velocity'', across frames varies from layer to layer.
In particular, deeper layers in the feature hierarchy change more slowly than shallower layers over video sequences.
We propose that network execution can be viewed as an aspect of architecture and define the ``clockwork'' FCN  (c.f. clockwork recurrent networks \cite{koutnik2014clockwork}).
Combining these two insights, we group the layers of the network into stages, and set separate update rates for these levels of representation.
The execution of a stage on a given frame is determined by either a fixed clock rate (``fixed-rate") or data-driven (``adaptive").
The prediction for the current frame is then the fusion (via the skip layer architecture of the FCN) of these computations on multiple frames, thus exploiting the lower resolution and slower rate-of-change of deeper layers.

We demonstrate the efficacy of the architecture for both fixed and adaptive schedules.
We show results on multiple datasets for a pipelining schedule designed to reduce latency for real-time recognition as well as a fixed-rate schedule designed to reduce computation and hence time and power.
Next we learn the clock-rate adaptively from the data, and demonstrate computational savings when little motion occurs in the video without sacrificing recognition accuracy during dynamic scenes.
We verify our approach on synthetic frame sequences made from PASCAL VOC~\cite{pascal} and evaluate on videos from the NYUDv2~\cite{nyud}, YouTube-Objects~\cite{prest2012learning}, and Cityscapes~\cite{cordts2016cityscapes} datasets.

\begin{figure}[t]
\begin{center}
   \includegraphics[width=0.95\linewidth]{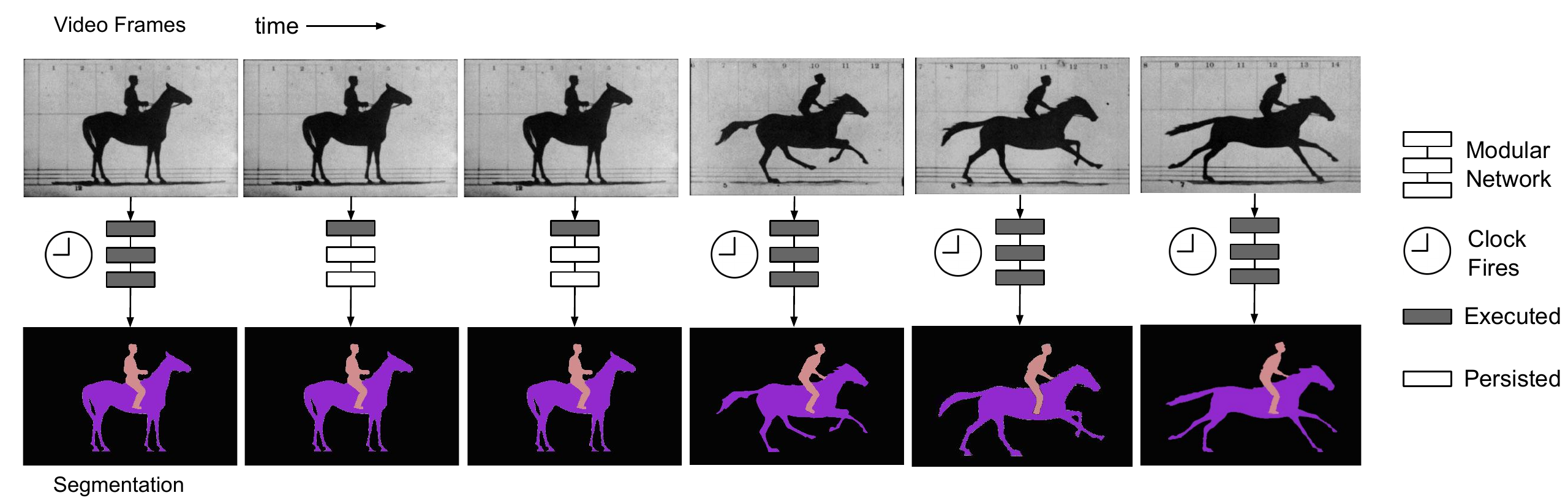}
\end{center}
\caption{
Our adaptive clockwork method illustrated with the famous \emph{The Horse in Motion}~\cite{muybridge}, captured by Eadweard Muybridge in 1878 at the Palo Alto racetrack.
The clock controls network execution: past the first stage, computation is scheduled only at the time points indicated by the clock symbol.
During static scenes cached representations persist, while during dynamic scenes new computations are scheduled and output is combined with cached representations.
}
\label{fig:overview}
\end{figure}

\section{Related Work}
\label{sec:related}
We extend fully convolutional networks for image semantic segmentation to video semantic segmentation.
Convnets have been applied to video to learn spatiotemporal representations for classification and detection but rarely for dense pixelwise, frame-by-frame inference.
Practicality requires network acceleration, but generic techniques do not exploit the structure of video.
There is a large body of work on video segmentation, but the focus has not been on \emph{semantic} segmentation, nor are methods computationally feasible beyond short video shots.

\minisection{Fully Convolutional Networks}
A fully convolutional network (FCN) is a model designed for pixelwise prediction \cite{fcn}.
Every layer in an FCN computes a local operation, such as convolution or pooling, on relative spatial coordinates.
This locality makes the network capable of handling inputs of any size while producing output of corresponding dimensions.
Efficiency is preserved by computing single, dense forward inference and backward learning passes.
Current classification architectures -- AlexNet \cite{alexnet}, GoogLeNet \cite{googlenet}, and VGG \cite{vgg} -- can be cast into corresponding fully convolutional forms.
These networks are learned end-to-end, are fast at inference and learning time, and can be generalized with respect to different image-to-image tasks.
FCNs yield state-of-the-art results for semantic segmentation \cite{fcn}, boundary prediction \cite{xie2015holistically}, and monocular depth estimation \cite{eigen2015predicting}.
While these tasks process each image in isolation, FCNs extend to video.
As more and more visual data is captured as video, the baseline efficiency of fully convolutional computation will not suffice.

\minisection{Video Networks and Frame Selection}
Time can be incorporated into a network by spatiotemporal filtering or recurrence.
Spatiotemporal filtering, i.e. 3D convolution, can capture motion for activity recognition \cite{ji2013conv3d,karpathy2014large}.
For video classification, networks can integrate over time by early, late, or slow fusion of frame features \cite{karpathy2014large}.
Recurrence can capture long-term dynamics and propagate state across time, as in the popular long short-term memory (LSTM) \cite{lstm}.
Joint convolutional-recurrent networks filter within frames and recur across frames: the long-term recurrent convolutional network \cite{lrcn} fuses frame features by LSTM for activity recognition and captioning.
Frame selection reduces computation by focusing computational resources on important frames identified by the model: space-time interest points \cite{laptev2005space} are video keypoints engineered for sparsity, and a whole frame selection and recognition policy can be learned end-to-end for activity detection \cite{yeung2015end}.
For optical flow, an intrinsically temporal task, a cross-frame FCN is state-of-the-art among fast methods \cite{fischer2015flownet}.
These video recognition approaches none address frame-by-frame output.

\minisection{Network Acceleration}
Although FCNs are fast, video demands computation that is faster still, particularly for real-time inference.
The spatially dense operation of the FCN amortizes the computation of overlapping receptive fields common to contemporary architectures.
However, the standard FCN does nothing to temporally amortize the computation of sequential inputs.
Computational concerns can drive architectural choices.
For instance, GoogLeNet requires less computation and memory than VGG, although its segmentation accuracy is worse \cite{fcn}.
Careful but time-consuming model search can improve networks within a fixed computational budget \cite{he2014convolutional}.
Methods to reduce computation and memory include reduced precision by weight quantization \cite{vanhoucke2011improving}, low-rank approximations with clustering, \cite{denton2014exploiting}, low-rank approximations with end-to-end tuning \cite{jaderberg2014speeding}, and kernel approximation methods like the fast food transformation \cite{yang2015deep}.
None of these generic acceleration techniques harness the frame-to-frame structure of video.
The proposed clockwork speed-up is orthogonal and compounds any reductions in absolute inference time.
Our clockwork insight holds for all layered architectures whatever the speed/quality operating point chosen.

\minisection{Semantic Segmentation}
Much work has been done to address the problem of segmentation in video.
However, the focus has not been on semantic segmentation.
Instead research has addressed spatio-temporal ``supervoxels'' \cite{grundmann2010efficient,xu2012evaluation}, unsupervised and motion-driven object segmentation \cite{shi1998motion,papazoglou2013fast,fragkiadaki2015learning}, or weakly supervising the segmentation of tagged videos \cite{hartmann2012weakly,tang2013discriminative,liu2014weakly}.
These methods are not suitable for real-time or the complex multi-class, multi-object scenes encountered in semantic segmentation settings.
Fast Object Segmentation in Unconstrained Videos \cite{papazoglou2013fast} infers only figure-ground segmentation at 0.5s/frame with offline computed optical flow and superpixels.
Although its proposals have high recall, even when perfectly parallelized \cite{fragkiadaki2015learning} this method takes $>15$s/frame and a separate recognition step is needed for semantic segmentation.
In contrast the standard FCN computes a full semantic segmentation in 0.1s/frame.

\section{Fast Frames and Slow Semantics}
\label{sec:velocities}
Our approach is inspired by observing the time course of learned, hierarchical features over video sequences.
Drawing on the local-to-global idea of skip connections for fusing global, deep layers with local, shallow layers, we reason that the semantic representation of deep layers is relevant across frames whereas the shallow layers vary with more local, volatile details.
Persisting these features over time can be seen as a temporal skip connection.

Measuring the relative difference of features across frames confirms the temporal coherence of deeper layers.
Consider a given score layer (a linear predictor of pixel class from features), $\ell$, with outputs $S_{\ell} \in [K \times H \times W]$, where $K$ is the number of categories and $H$, $W$ is the output dimensions for layer $\ell$.
We can compute the difference at time $t$ with a score map distance function $d_{\text{sm}}$, chosen to be the hamming distance of one hot encodings.
$$
d_{\text{sm}}(S_{\ell}^{t}, S_{\ell}^{t-1}) = d_{\text{hamming}}(\phi(S_{\ell}^t), \phi(S_{\ell}^{t-1}))
$$

Table~\ref{table:td} reports the average of these temporal differences for the score layers, as computed over all videos in the YouTube-Objects dataset~\cite{prest2012learning}.
It is perhaps unsurprising that the deepest score layer changes an order of magnitude less than the shallower layers on average.
We therefore hypothesize that caching deeper layer scores from past frames can inform the inference of the current frame with relatively little reduction in accuracy.

The slower rate of change of deep layers can be attributed to architectural and learned invariances.
More pooling affords more robustness to translation and noise, and learned features may be tuned to the supervised classes instead of general appearance.

\setlength{\tabcolsep}{12pt}
\begin{table}[h]
  \begin{center}
  \small{
  \begin{tabular}{c c c c}
    \toprule
    score layer & temporal difference & depth & semantic accuracy \\
    \midrule
    \texttt{pixels}& .26 $\pm$ .18 & 0  & -  \\
    \texttt{pool3} & .11 $\pm$ .06 & 9  & 9.6\%  \\
    \texttt{pool4} & .11 $\pm$ .06 & 13 & 20.7\% \\
    \texttt{fc7}   & .02 $\pm$ .02 & 19 & 65.5\% \\
    \bottomrule
  \end{tabular}
  }
  \end{center}
  \caption{
    The average temporal difference over all YouTube-Objects videos of the respective pixelwise class score outputs from a spectrum of network layers.
    The deeper layers are more stable across frames -- that is, we observe supervised convnet features to be ``slow'' features \cite{wiskott2002slow}.
    The temporal difference is measured as the proportion of label changes in the output.
    The layer depth counts the distance from the input in the number of parametric and non-linear layers.
    Semantic accuracy is the intersection-over-union metric on PASCAL VOC of our frame processing network fine-tuned for separate output predictions (Section \ref{sec:results}).
  }
  \label{table:td}
\end{table}
\setlength{\tabcolsep}{6pt}

\begin{figure}[h]
\centering
\includegraphics[width=.48\linewidth]{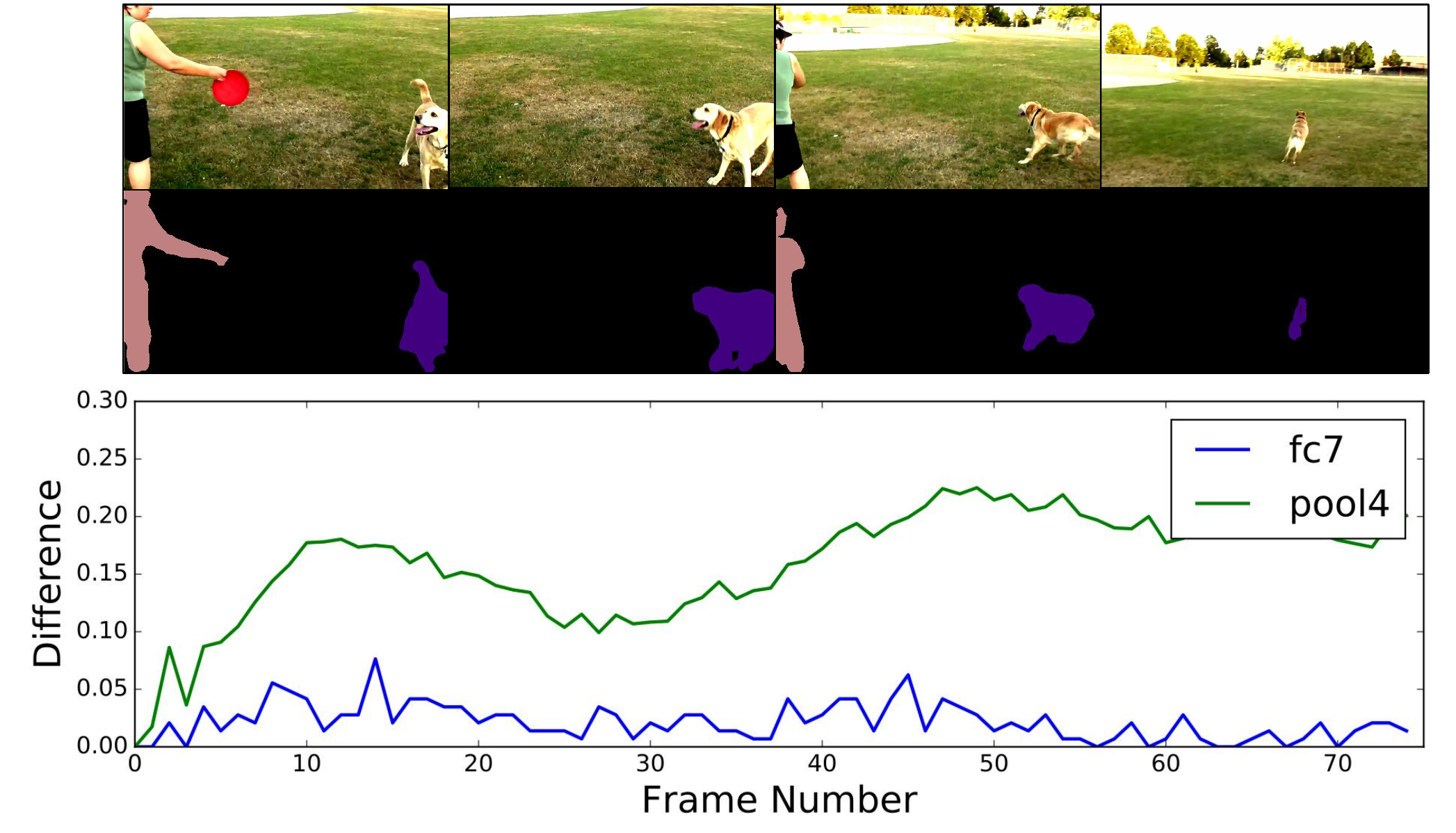}
\includegraphics[width=.48\linewidth]{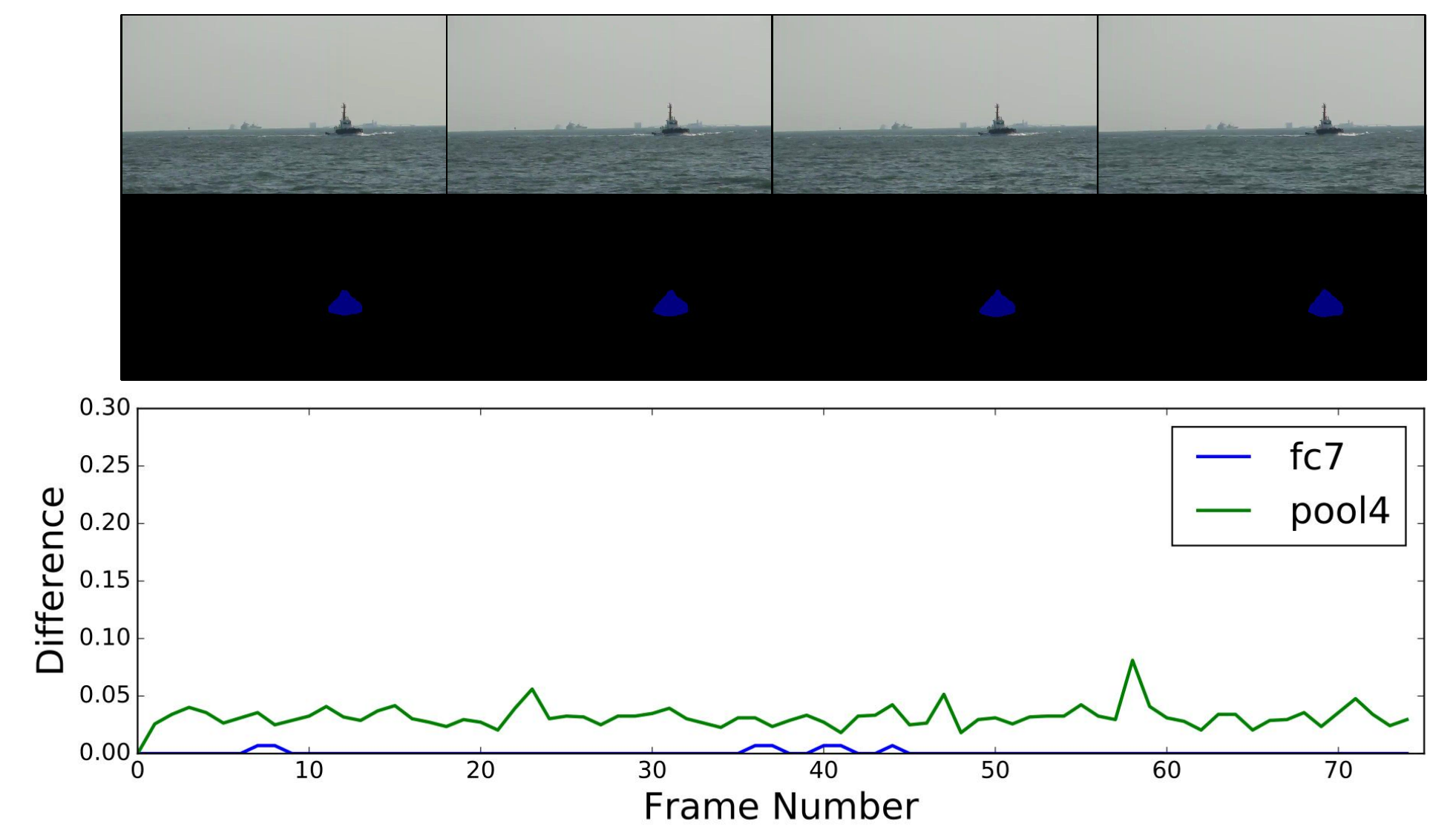}
\caption{
The proportional difference between adjacent frames of semantic predictions from a mid-level layer (\texttt{pool4}, green) and the deepest layer (\texttt{fc7}, blue) are shown for the first 75 frames of two videos.
We see that for a video with lots of motion (left) the difference values are large while for a relatively static video (right) the difference values are small.
In both cases, the differences of the deeper \texttt{fc7} are smaller than the differences of the shallower \texttt{pool4}.
The ``velocity'' of deep features is slow relative to shallow features and most of all the input.
At the same time, the differences between shallow and deep layers are dependent since the features are compositional.
}
\label{fig:sm_diff}
\end{figure}

While deeper layers are more stable than shallower layers, for videos with enough motion the score maps throughout the network may change substantially.
For example, in Figure~\ref{fig:sm_diff} we show the differences for the first 75 frames of a video with large motion (left) and with small motion (right).
We would like our network to adaptively update only when the deepest, most semantic layer (\texttt{fc7}) score map is likely to change.
We notice that though the intermediate layer (\texttt{pool4}) difference is always larger than the deepest layer difference for any given frame, the \texttt{pool4} differences are much larger for the video with large motion than for the video with relatively small motion.
This observation forms the motivation for using the intermediate differences as an indicator to determine the firing of an adaptive clock.

\section{A Clockwork Network}
\label{sec:clockwork}
We adapt the fully convolutional network (FCN) approach for image-to-image mapping \cite{fcn} to video frame processing.
While it is straightforward to perform inference with a still-image segmentation network on every video frame, this na\"ive computation is inefficient.
Furthermore, disregarding the sequential nature of the input not only sacrifices efficiency but discards potential temporal recognition cues.
The temporal coherence of video suggests the persistence of visual features from prior frames to inform inference on the current frame.
To this end we define the clockwork FCN, inspired by the clockwork recurrent network \cite{koutnik2014clockwork}, to carry temporal information across frames.
A generalized notion of clockwork relates both of these networks.

We consider both throughput and latency in the execution of deep networks across video sequences.
The inference time of the regular FCN-8s at $\sim100$ms per frame of size $500 \times 500$ on a standard GPU can be too slow for video.  
We first define fixed clocks then extend to adaptive and potentially learned clockwork to drive network processing.
\begin{figure}[h]
\begin{center}
   \includegraphics[width=0.95\linewidth]{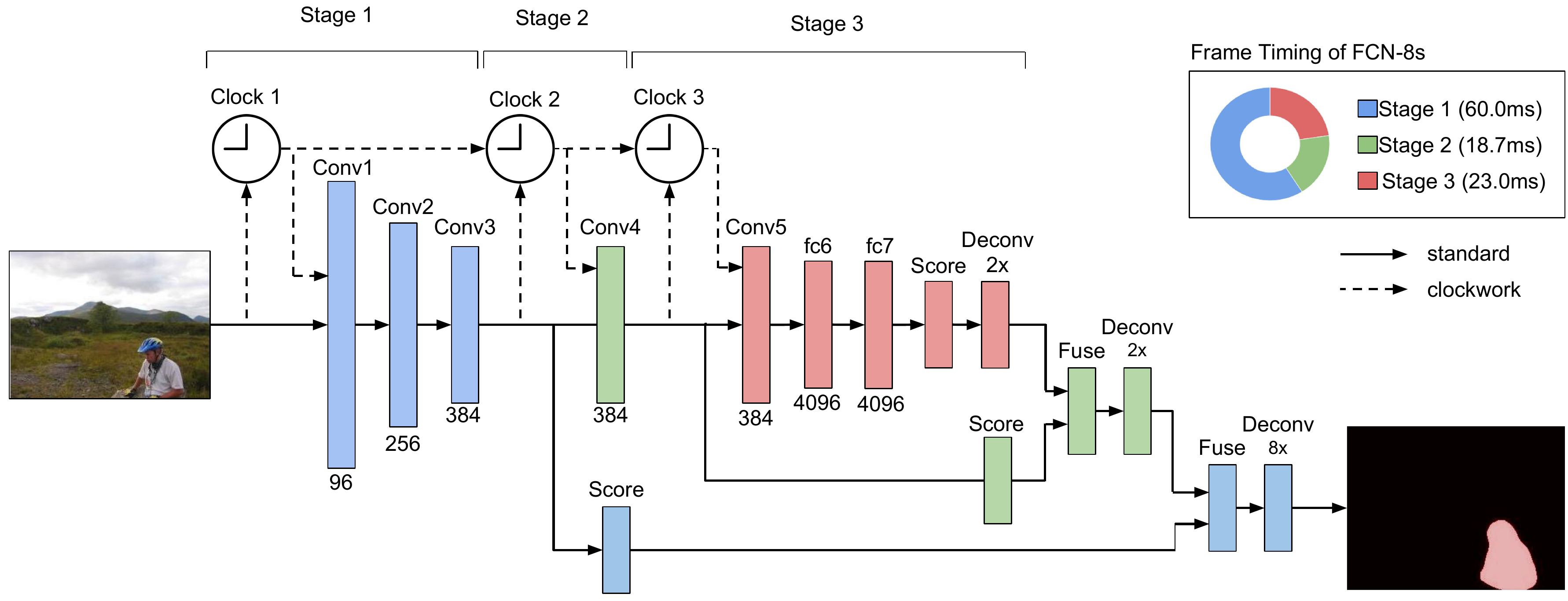}
\end{center}
\caption{
The clockwork FCN with its stages and corresponding clocks.
}
\label{fig:clockfcn}
\end{figure}
Whatever the task, any video network can be accelerated by our clockwork technique.
A schematic of our clockwork FCN is shown in Figure \ref{fig:clockfcn}.

\begin{figure}[t]
\begin{center}
   \includegraphics[width=0.95\linewidth]{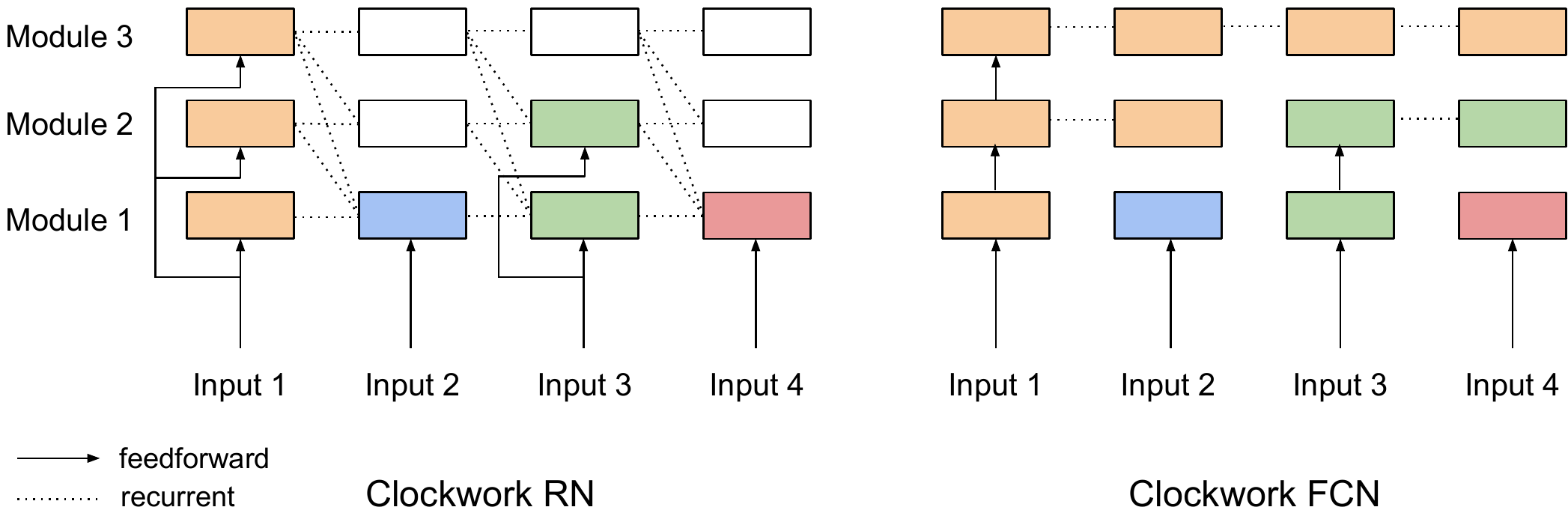}
\end{center}
\caption{
A comparison of the layer connectivity and time course of outputs in the clockwork recurrent network \cite{koutnik2014clockwork} and in our clockwork FCN.
Module color marks the time step of evaluation, and blank modules are disconnected from the network output.
The clock RN is flat with respect to the input while our network has a hierarchical feature representation.
Each clock RN module is temporally connected to itself and slower modules while in our network each module is only temporally connected to itself.
Features persist over time in both architectures, but in our architecture they contribute to the network output at each step.
}
\label{fig:clockfeat}
\end{figure}

There are several choice points in defining a clockwork architecture.
We define a novel, generalized clockwork framework, which can purposely schedule deeper layers more slowly than shallower layers.
We form our modules by grouping the layers of a convnet to span the feature hierarchy.
Our networks persists both state and output across time steps.
The clockwork recurrent network of \cite{koutnik2014clockwork}, designed for long-term dependency modeling of time series, is an instance of our more general scheme for clockwork computation.
The differences in architecture and outputs over time between clockwork recurrence and our clockwork are shown in Figure \ref{fig:clockfeat}.

While different, these nets can be expressed by generalized clockwork equations
\begin{align}
  y_H^{(t)} &= f_T\left( C_H^{(t)} \elprod f_H(y_H^{(t-1)}) + C_I^{(t)} \elprod f_I(x^{(t)}) \right) \label{eq:hidden} \\
  y_O^{(t)} &= f_O\left( C_O^{(t)} \elprod f_H(y_H^{(t)})\right) \label{eq:output}
\end{align}
with the state update defined by Equation \ref{eq:hidden} and the output defined by Equation \ref{eq:output}.
The  data $x^{(t)}$, hidden state $y_H^{(t)}$ output $y_O^{(t)}$ vary with time $t$.
The functions $f_I, f_H, f_O, f_T$ define input, hidden state, output, and transition operations respectively and are fixed across time.
The input, hidden, and output clocks $C_I^{(t)}, C_H^{(t)}, C_O^{(t)}$ modulate network operations by the elementwise product $\elprod$ with the corresponding function evaluations.
We recover the standard recurrent network (SRN), clockwork recurrent network (clock RN), and our network (clock FCN) in this family of equations.
The settings of functions and clocks are collected in Table \ref{table:clockeq}.

\begin{table}[h]
  \begin{center}
  \small{
  \begin{tabular}{c c c c c c c c}
    \toprule
    network & $f_I$ & $f_H$ & $f_O$ & $f_T$ & $C_I$ & $C_H$ & $C_O$ \\
    \midrule
    SRN       & $W_I$   & $W_H$ & TanH & TanH & $\ones$ & $\ones$         & $\ones$ \\
    clock RN  & $W_I$   & $W_H$ & TanH & TanH & $C$     & $C$             & $C$ \\
    clock FCN & $\circ$ & $I$   & ReLU & $I$  & $C$     & $\overline{C} $ & $\ones$ \\
    \bottomrule
  \end{tabular}
  }
  \end{center}
  \caption{
    The standard recurrent network (SRN), clockwork recurrent network (clock RN), and our network (clock FCN) in generalized clockwork form.
    The recurrent networks have learned hidden weights $W_H$ and non-linear transition functions $f_T$, while clock FCN persists state by the identity $I$.
    Both recurrent modules are flat with linear input weights $W_I$, while clock FCN modules have hierarchical features by layer composition $\circ$.
    The SRN has trivial constant, all-ones $\ones$ clocks.
    The clock RN has a shared input, hidden, and output clock with exponential rates.
    Our clock FCN has alternating input and hidden clocks $C, \overline{C}$ to compute or cache and has a constant, all-ones $\ones$ output clock to fuse output on every frame.
  }
  \label{table:clockeq}
\end{table}

Inspired by the clockwork RN, we investigate persisting features and scheduling layers to process video with a semantic segmentation convnet.
Recalling the lessened semantic rate of deeper layers observed in Section \ref{sec:velocities}, the skip layers in FCNs originally included to preserve resolution by fusing outputs are repurposed for this staged computation.
We cache features and outputs over time at each step to harness the continuity of video.
In contrast, the clockwork RN persists state but output is only made according to the clock, and each clockwork RN module is connected to itself and all slower modules across time whereas a module in our network is only connected to itself across time.

\subsection{Execution as Architecture}
Clockwork architectures partition a network into modules or stages that are executed according to different schedules.
In the standard view, the execution of an architecture is an all-or-nothing operation that follows from the definition of the network.
Relaxing the strict identification of architecture and execution instead opens up a range of potential schedules.
These schedules can be encompassed by the introduction of one first-class architectural element: the clock.

A clock defines a dynamic cut in the computation graph of a network.
As clocks mask state in the representation, as detailed in  Equations \ref{eq:hidden} and \ref{eq:output}, clocks likewise mask execution in the computation.
When a clock is on, its edges are intact and execution traverses to the next nodes/modules.
When a clock is off, its edges are cut and execution is blocked.
Alternatives such as computing the next stage or caching a past stage can be scheduled by a paired clock $C$ and counter-clock $\overline{C}$ with complementary sets of edges.
Any layer (or composition of layers) with binary output can serve as a clock.
As a layer, a clock can be fixed or learned.
For instance, the following are simple clocks of the form $f(x, t)$ for features $x$ and time $t$:
\begin{itemize}
  \item $1$ to always execute
  \item $t \equiv 0 \pmod{2}$ to execute every other time
  \item $\|x_{t} - x_{t-1} \| > \theta$ to execute for a difference threshold
\end{itemize}

\subsection{Networks in Time}

Having incorporated scheduling into the network with clocks, we can optimize the schedule for various tasks by altering the clocks.

\minisection{Pipelining}
To reduce latency for real-time recognition we pipeline the computation of sequential frames analogously to instruction pipelining in processors.
We instantiate a three-stage pipeline, in which stage 1 reflects frame $i$, stage 2 frame $i-1$, and stage 3 frame $i-2$.
The total time to process the frame is the time of the longest stage, stage $1$ in our pipeline, plus the time for interpolating and fusing outputs.
Our 3-stage pipeline FCN reduces latency by $59\%$.
A 2-stage variation further balances latency and accuracy.

\minisection{Fixed-Rate}
To reduce overall computation we limit the execution rates of stages and persist features across frames for skipped stages.
Given the learned invariance and slow semantics of deep layers observed in Section \ref{sec:velocities}, the deeper layers can be executed at a lower rate to save computation while other stages update.
These clock rates are free parameters in the schedule for exchanging inference speed and accuracy.
We again divide the network into three stages, and compare rates for the stages.
The exponential clockwork schedule is the natural choice of halving the rate at each stage for more efficiency.
The alternating clockwork schedule consolidates the earlier stages to execute these on every frame and executes the last stage on every other frame for more accuracy.
These different sets of rates cover part of the accuracy/efficiency spectrum.

The current stages are divided into the original score paths of the FCN-8s architecture, but they need not be.
One could prioritize latency, spatial refinement, or certain output classes by rebalancing the computation.
It is possible to partially compute a span of layers and defer their full execution to a following stage; this can be accomplished by sparse evaluation through dynamic striding and dilation \cite{yu2015dilate}.
In principle the stage progression can be decided online in lieu of fixing a schedule for all inference.
We turn to adaptive clockwork for deciding execution.

\subsection{Adaptive Clockwork}
\label{sub:clockadaptive}
All of the clocks considered thus far have been fixed functions of time but not the data.
Setting these clocks gives rise to many schedules that can be tuned to a given task or video, but this introduces a tedious dimension of model search.
Much of the video captured in the wild is static and dynamic in turn with a variable amount of motion and semantic progression at any given time.
Choosing many stages or a slow clock rate may reduce computation, but will likewise result in a steep decline in accuracy for dynamic scenes.
Conversely, faster update rates or fewer stages may capture transitory details but will needlessly compute and re-compute stable scenes.
Adaptive clocks fire based on the input and network state, resulting in a responsive schedule that varies with the dynamism of the scene.
The clock can fire according to any function of the input and network state.
A difference clock can fire on the temporal difference of a feature across frames.
A confidence clock can fire on peaks in the score map for a single frame.
This approach extends inference from a pre-determined architecture to a set of architectures to choose from for each frame, relying on the full FCN for high accuracy in dynamic scenes while taking advantage of cached representations in more static scenes.
\begin{align*}
\text{threshold clock} \; & \|x_{t} - x_{t-1} \| > \theta & \text{learned clock} \; & f_\theta(x_{t}, x_{t-1})
\end{align*}
The simplest adaptive clock is a threshold, but adaptive clocks could likewise be learned (for example as a temporal convolution across frames).
The threshold can be optimized for a specific tradeoff along the accuracy/efficiency curve.
Given the hierarchical dependencies of layers and the relative stability of deep features observed in Section \ref{sec:velocities}, we threshold differences at a shallower stage for adaptive scheduling of deeper stages.
The sensitivity of the adaptive clock can even be set on unannotated video by thresholding the proportional temporal difference of output labels as in Table \ref{table:td}.
Refer to Section \ref{sub:adaptive} for the results of threshold-adaptive clockwork with regard to clock rate and accuracy.

\section{Results}
\label{sec:results}
Our base network is FCN-8s, the fully convolutional network of \cite{fcn}.
The architecture is adapted from the VGG16 architecture~\cite{vgg} and fine-tuned from ILSVRC pre-training.
The net is trained with batch size one, high momentum, and all skip layers at once.

In our experiments we report two common metrics for semantic segmentation that measure the region intersection over union (IU):

\begin{itemize}
  \item mean IU: \scalebox{1.0}{$(1/n_{\text{cl}}) \sum_i n_{ii} /\left(t_i + \sum_j n_{ji} - n_{ii}\right)$} 
  \item frequency weighted IU: \scalebox{1.0}{$\left(\sum_k t_k\right)^{-1} \sum_i t_i n_{ii} /\left(t_i + \sum_j n_{ji} - n_{ii}\right)$} 
\end{itemize}
for $n_{ij}$ the number of pixels of class $i$ predicted to belong to class $j$, where there are $n_{\text{cl}}$ different classes, and for $t_i = \sum_j n_{ij}$ the total number of pixels of class $i$.

\noindent We evaluate our clockwork FCN on four video semantic segmentation datasets.

\minisection{Synthetic sequences of translated scenes}
We first validate our method by evaluating on synthetic videos of moving crops of PASCAL VOC images~\cite{pascal} in order to score on a ground truth annotation at every frame.
For source data, we select the 736 image subset of the PASCAL VOC 2011 segmentation validation set used for FCN-8s validation in \cite{fcn}.
Video frames are generated by sliding a crop window across the image by a predetermined number of pixels, and generated translations are vertical or horizontal according to the portrait or landscape aspect of the chosen image.
Each synthetic video is six frames long.
For each seed image, a ``fast'' and ``slow'' video is made with 32 pixel and 16 pixel frame-to-frame displacements respectively.

\minisection{NYU-RGB clips}
The NYUDv2 dataset~\cite{nyud} collects short RGB-D clips and includes a segmentation benchmark with high-quality but temporally sparse pixel annotations (every tenth video frame is labeled).
We run on video from the ``raw" clips subsampled 10X and evaluate on every labeled frame.
We consider RGB input alone as the depth frames of the full clips are noisy and uncurated.
Our pipelined and fixed-rate clockwork FCNs are run on the entire clips and accuracy is reported for those frames included in the segmentation test set.

\minisection{Youtube-Objects}
The Youtube-Objects dataset~\cite{prest2012learning} provides videos collected from Youtube that contain objects from ten PASCAL classes.
We restrict our attention to a subset of the videos that have pixelwise annotations by \cite{jainsupervoxel} as the original annotations include only initial frame bounding boxes.
This subset was drawn from all object classes, and contains 10,167 frames from 126 shots, for which every 10th frame is human-annotated.
We run on only annotated frames, effectively 10X subsampling the video.
We directly apply our networks derived from PASCAL VOC supervision and do not fine-tune to the video annotations.

\minisection{Cityscapes}
The Cityscapes dataset~\cite{cordts2016cityscapes} collects frames from video recorded at 17hz by a car-mounted camera while driving through German cities.
While annotations are temporally sparse, the preceding and following input frames are provided.
Our network is learned on the \texttt{train} split and then all schedules are evaluated on \texttt{val}.

\subsection{Pipelining}
\label{sec:pipeline}
\begin{table}[h]
\centering
\resizebox{.9\columnwidth}{!}{
\begin{tabular}{l c c c c c}
\toprule
16 pixel shift & Time (\% of full)& Mean IU & fwIU & Mean IU-bdry & fwIU-bdry \\
\midrule
3-Stage Baseline & 59\% & 9.2     & 52.6    & 6.1     & 9.4      \\
3-Stage Pipeline & 59\%  & 56.0    & 76.5    & 44.6    & 42.9 \\
2-Stage Baseline & 77\%  & 22.5    & 64.7    & 16.6    & 21.9      \\
2-Stage Pipeline & 77\%  & \bf63.3 & \bf81.7 & \bf52.3 & \bf51.0 \\
\midrule
Frame Oracle & 100\% & 65.9 & 83.6 & 57.0 & 56.3 \\
\bottomrule
\toprule
32 pixel shift   & Time (\% of full) & Mean IU   & fwIU  & Mean IU-bdry & fwIU-bdry \\
\midrule
3-Stage Baseline & 59\% & 9.2       & 52.6     & 6.0     & 9.4      \\
3-Stage Pipeline & 59\% & 45.5      & 67.4     & 37.7    & 36.0 \\
2-Stage Baseline & 77\% & 22.4      & 62.8     & 16.2    & 21.7     \\
2-Stage Pipeline & 77\% & \bf  57.8 & \bf 76.6 & \bf46.6 & \bf45.1\\
\midrule
Frame Oracle     & 100\%            & 65.6     & 82.6    & 55.8         & 55.3 \\
\bottomrule
\end{tabular}
}
\caption{\small
  Pipelined segmentation of translated PASCAL sequences.
  Synthesized video of translating PASCAL scenes allows for assessment of the pipeline at every frame.
  The pipelined FCN segments with higher accuracy in the same time envelope as the every-other-frame evaluation of the full FCN.
  Metrics are computed on the standard masks and a 10-pixel band at boundaries.
}
\label{table:pipepascal}
\end{table}

Pipelined execution schedules reduce latency by producing an output each time the first stage is computed. Later stages are persisted from previous frames and their outputs are fused with the output of the first stage computed on the current frame.
The number of stages is determined by the number of clocks.
We consider a full \textbf{3-stage pipeline} and a condensed \textbf{2-stage pipeline} where the stages are defined by the modules in Figure \ref{fig:clockfcn}.
In the pipelined schedule, all clock rates are set to 1, but clocks fire simultaneously to update every stage in parallel.
This is made possible by asynchrony in stage state, so that a later stage is independent of the current frame but not past frames.

To assess our pipelined accuracy and speed, we compare to reference methods that bound both recognition and time.
A frame oracle evaluates the full FCN on every frame to give the best achievable accuracy for the network independent of timing.
As latency baselines for our pipelines, we truncate the FCN to end at the given stage.
Both of our staged, pipelined schedules execute at lower latency than the oracle with better accuracy for fixed latency than the baselines.
We verify these results on synthetic PASCAL sequences as reported in Table \ref{table:pipepascal}.
Results on PASCAL, NYUD, and YouTube are reported in Table \ref{table:pipefull}.

Our pipeline scheduled networks reduce latency with minimal accuracy loss relative to the standard FCN run on each frame without time restriction.
These quantitative results demonstrate that the deeper layer representations from previous frames contain useful information that can be effectively combined with low-level predictions for the current frame.

\begin{table}[h]
\centering
\resizebox{\columnwidth}{!}{
\begin{tabular}{l c c c c c c c c c}
\toprule
  & & \multicolumn{2}{c}{NYUD} && \multicolumn{2}{c}{Youtube} && \multicolumn{2}{c}{Pascal Shift 16}\\
  \cline{3-4} \cline{6-7} \cline{9-10}\\
  Schedule & Time (\% of full) & Mean IU & fwIU & & Mean IU & fwIU & & Mean IU & fwIU\\
  \midrule
  3-Stage Baseline & 59\% & 8.1     & 22.2    && 12.2     & 74.2    && 9.2     & 54.7     \\
  3-Stage Pipeline & 59\% & 25.1    & 38.0    && 58.1     & 87.0    && 56.0    & 76.5 \\
  2-Stage Baseline & 77\% & 16.5    & 32.1    && 21.5     & 7.8     && 22.5    & 64.7    \\
  2-Stage Pipeline & 77\% & \bf26.4 & \bf39.5 && \bf 64.0 & \bf89.2 && \bf63.3 & \bf81.7  \\
  \midrule
  Frame Oracle     & 100\% & 31.1 & 45.5 && 70.0 & 91.5    && 65.9    & 83.6 \\
\bottomrule
\end{tabular}
}
\caption{\small
  Pipelined execution of semantic segmentation on three different datasets.
  Inference approaches include pipelines of different lengths and a full FCN frame oracle.
  We also show baselines with comparable latency to the pipeline architectures.
  Our pipelined network offers the best accuracy of computationally comparable approaches running near frame rate.
  The loss in accuracy relative to the frame oracle is less than the relative speed-up.
}
\label{table:pipefull}
\end{table}

\begin{figure*}[h!]
  \centering
  \includegraphics[width=\linewidth]{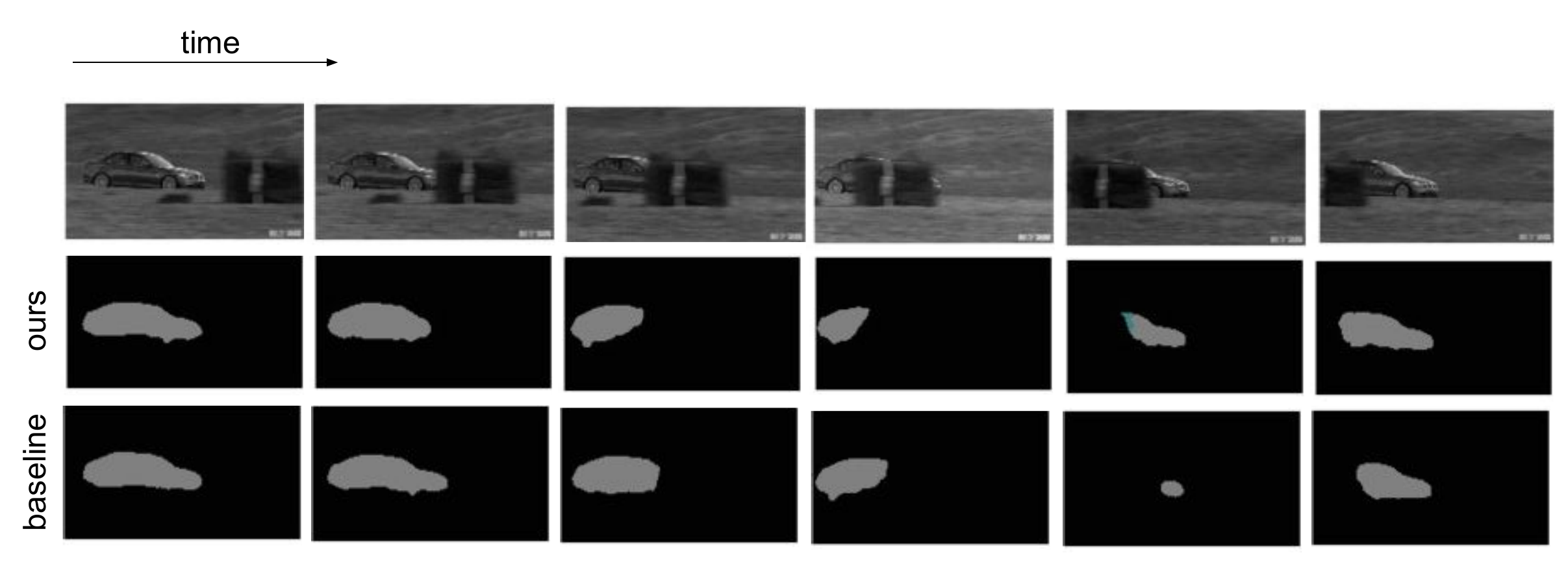}
  \caption{Pipelined vs. standard FCN on YouTube video. Our method is able to detect the occlusion of the car as it is happening unlike the lagging baseline computed on every other frame.}
  \label{fig:youtube}
\end{figure*}

We show a qualitative result for our pipelined FCN on a sequence from the YouTube-Objects dataset~\cite{prest2012learning}.
Figure~\ref{fig:youtube} shows one example where our pipeline FCN is particularly useful.
Our network quickly detects the occlusion of the car while the baseline lags and does not immediately recognize the occlusion or reappearance.

\setlength{\floatsep}{4pt}
\setlength{\intextsep}{12pt}
\setlength{\textfloatsep}{12pt}
\setlength{\floatsep}{\the\floatsep}
\setlength{\intextsep}{\the\intextsep}
\setlength{\textfloatsep}{\the\textfloatsep}

\subsection{Fixed-Rate}
Fixed-rate clock schedules reduce overall computation relative to full, every frame evaluation by assigning different update rates to each stage such that later stages are executed less often.
Rates can be set aggressively low for extreme efficiency or conservatively high to maintain accuracy while sparing computation.
The \textbf{exponential clockwork} schedule executes the first stage on every frame then updates following stages exponentially less often by halving with each stage.
The \textbf{alternating clockwork} schedule combines stages 2 and 3, executes the first stage on every frame, then schedules the following combined stage every other frame.

A frame oracle that evaluates the full FCN on \textit{every} frame is the reference model for accuracy.
Evaluating the full FCN on \textit{every other} frame is the reference model for computation.
Due to the distribution of execution time over stages, this is faster than either clockwork schedule, though clockwork offers higher accuracy.
Alternating clockwork achieves higher accuracy than the every other frame reference.
See Table \ref{table:clockpascal}.

\setlength{\intextsep}{12pt}
\setlength{\textfloatsep}{12pt}
\begin{table}[h]
\centering
\resizebox{.7\columnwidth}{!}{
\begin{tabular}{l c c c c c}
\toprule
16 pixel shift      & Clock Rates & Mean IU & fwIU    & Mean IU-bdry & fwIU-bdry \\
\midrule
Skip Frame Baseline & (2,2,2)     & 63.0    & 81.5    & 60.2         & 52.2 \\
Exponential         & (1,2,4)     & 61.4    & 80.4    & 50.5         & 49.1 \\
Alternating         & (1,1,2)     & \bf64.7 & \bf82.6 & \bf54.8      & \bf53.7 \\
\midrule
Frame Oracle        & (1,1,1)     & 65.9    & 83.6    & 57.0         & 56.3 \\
\bottomrule
\end{tabular}
}
\hfill%
\resizebox{.7\columnwidth}{!}{
\begin{tabular}{l c c c c c}
\toprule
32 pixel shift      & Clock Rates & Mean IU   & fwIU     & Mean IU-bdry & fwIU-bdry \\
\midrule
Skip Frame Baseline & (2,2,2)     & 59.5      & 77.9     & 49.4         & 48.2 \\
Exponential         & (1,2,4)     & 55.5      & 74.7     & 46.3         & 44.8 \\
Alternating         & (1,1,2)     & \bf  61.9 & \bf 79.6 & \bf51.7      & \bf50.6\\
\midrule
Frame Oracle        & (1,1,1)     & 65.6      & 82.6     & 55.8         & 55.3 \\
\bottomrule
\end{tabular}
}
\caption{\small
  Fixed-rate segmentation of translated PASCAL sequences.
  We evaluate the network on synthesized video of translating PASCAL scenes to assess the effect of persisting layer features across frames.
  Metrics are computed on the standard masks and a 10-pixel band at boundaries.
}
\label{table:clockpascal}
\end{table}

\begin{table}[h]
\centering
\resizebox{.8\columnwidth}{!}{
\begin{tabular}{l c c c c c c c c c}
\toprule
  & \multicolumn{2}{c}{NYUD} && \multicolumn{2}{c}{Youtube} && \multicolumn{2}{c}{Cityscapes}\\
  \cline{2-3} \cline{5-6} \cline{8-9}\\
  Schedule            & Mean IU & fwIU & & Mean IU & fwIU & & Mean IU & fwIU\\
\midrule
  Skip Frame Baseline & 27.7     & 41.3     && 65.6    & 89.7    && 62.1    & 87.4 \\
  Alternating         &  28.5 & 42.4 && 67.0    & 90.3    && \bf64.4 & \bf88.6  \\
  Adaptive            & \bf28.9  & \bf43.3 && \bf68.5 & \bf91.0 && 61.8    & 87.6 \\  
\midrule
  Frame Oracle        & 31.1 & 45.5 && 70.0 & 91.4  && 65.9 & 83.6 \\
\bottomrule
\end{tabular}
}
\caption{\small
  Fixed-rate and adaptive clockwork FCN evaluation. We score our network on
  three datasets with an alternating schedule that executes the later stage
  every other frame and an adaptive schedule that executes according to a
  frame-by-frame threshold on the difference in output. The adaptative
  threshold is tuned to execute the full network on $50\%$ of frames to
  equalize computation between the alternating and adaptive schedules.
}
\label{table:clockfull}
\end{table}

\setlength{\intextsep}{\the\intextsep}
\setlength{\textfloatsep}{\the\textfloatsep}

Exponential clockwork shows degraded accuracy yet takes $1.5\times$ the computation of evaluation on every other frame, so we discard this fixed schedule in favor of adaptive clockwork.
Although exponential rates suffice for the time series modeled by the clockwork recurrent network \cite{koutnik2014clockwork}, these rates deliver unsatisfactory results for the task of video semantic segmentation.
See Table \ref{table:clockfull} for alternating clockwork results on NYUD, YouTube-Objects, and Cityscapes.

\subsection{Adaptive Clockwork}
\label{sub:adaptive}
The best clock schedule can be data-dependent and unknown before segmenting a video.
Therefore, we next evaluate our adaptive clock rate as described in Section~\ref{sub:clockadaptive}.
In this case the adaptive clock only fully processes a frame if the relative difference in \texttt{pool4} score is larger than some threshold $\theta$.
This threshold may be interpreted as the the fraction of the score map that must switch labels before the clock updates the upper layers of the network.
See Table \ref{table:clockfull} for adaptive clockwork results on NYUD, YouTube-Objects, and Cityscapes.

We experiment with varying thresholds on the Youtube-Objects dataset to measure accuracy and efficiency.
We pick thresholds in $\theta=[0.1,0.5]$ as well as $\theta=0.0$ for unconditionally updating on every frame.
\begin{figure}
\begin{minipage}[t]{.5\linewidth}\centering
  \vspace{0pt}
  \includegraphics[width=\linewidth]{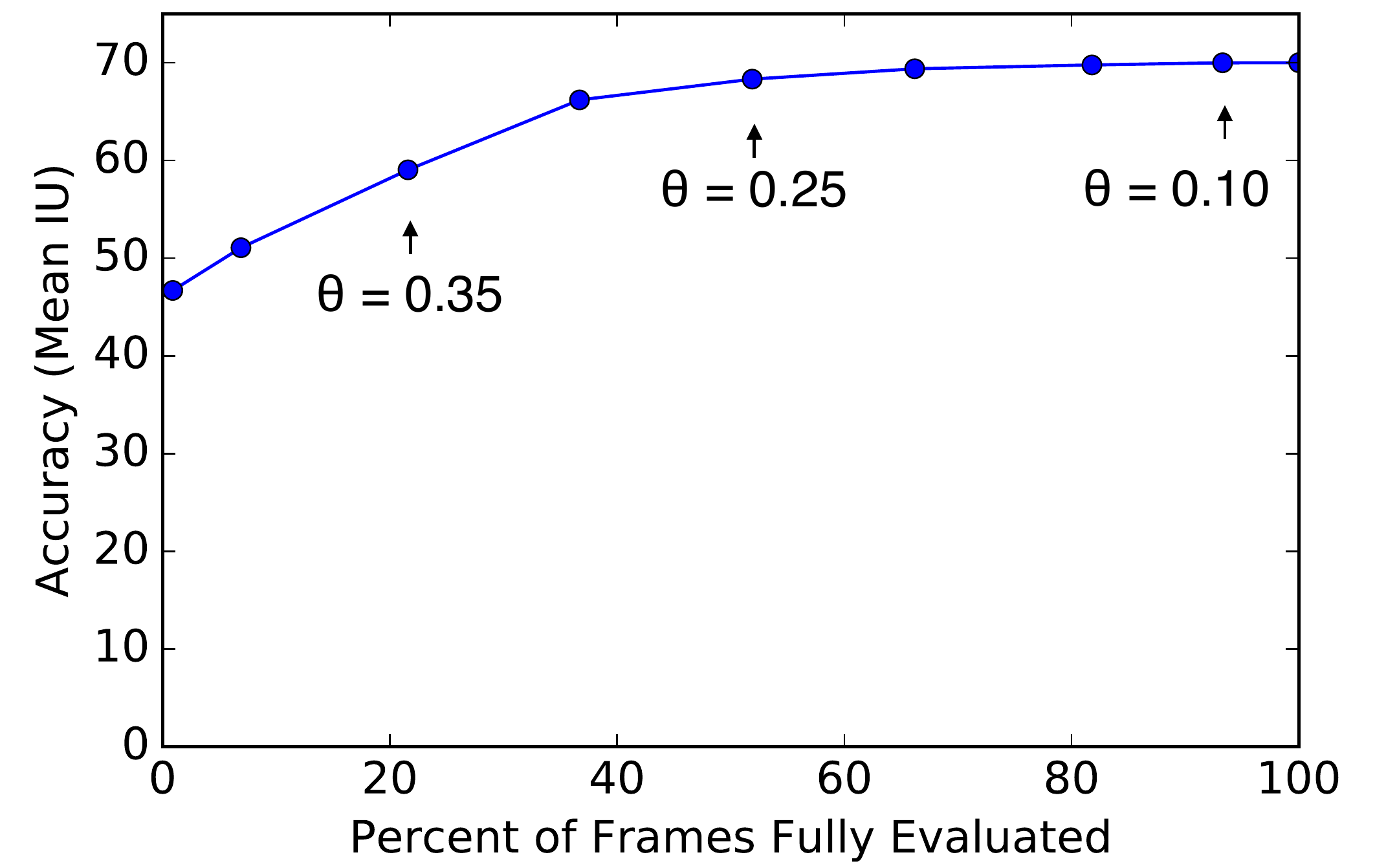}
\end{minipage}
\begin{minipage}[t]{.5\linewidth}\centering
    \vspace{.66cm}
  \resizebox{\linewidth}{!}{
  \begin{tabular}{l c c c c c c c c c c}
    \toprule
    Method  & \% Full Frames & Mean IU \\
    \midrule
    Adaptive [$\theta=0.10$] & 93\%  & 70.0 \\
    Adaptive [$\theta=0.25$] & 52\%  & 68.3 \\
    Adaptive [$\theta=0.35$] & 21\%  & 59.0 \\
    \midrule
    Frame Oracle             & 100\% & 70.0 \\
    \bottomrule
  \end{tabular}
  }
\end{minipage}

\caption{
Adaptive Clockwork performance across the Youtube-Objects dataset.
We examine various adaptive difference thresholds $\theta$ and plot accuracy (mean IU) against the percentage of frames that the adaptive clock chooses to fully compute.
A few corresponding thresholds are indicated.
}
  \label{fig:adapt-clk}
\end{figure}

In Figure~\ref{fig:adapt-clk} (left) we report mean IU accuracy as a function of our adaptive clock firing rate; that is, the percentage of frames the clock decides to fully process in the network.
The thresholds which correspond to a few points on this curve are indicated with mean IU (right).
Notice that our adaptive clockwork is able to fully process only 52\% of the frames while suffering a minimal loss in mean IU ($\theta=0.25$).
This indicates that our adaptive clockwork is capable of discovering semantically stationary scenes and saves significant computation by only updating when the output score map is predicted to change.

\setlength{\floatsep}{12pt}
\setlength{\intextsep}{12pt}
\setlength{\textfloatsep}{12pt}
\begin{figure}[h!]
\centering
\includegraphics[width=\linewidth]{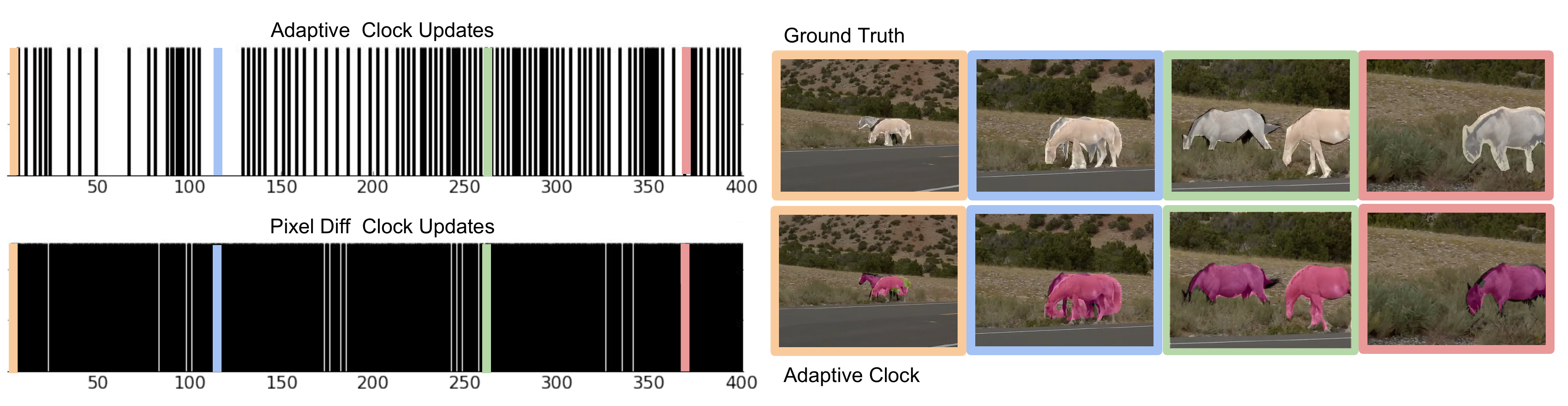}
\caption{
An illustrative example of our adaptive clockwork method on a video from Youtube-Objects.
On the left, we compare clock updates over time (shown in black) of our adaptive clock as well as a clock based on pixel differences.
Our adaptive clock updates the full network on only 26\% of the frames, determined by the threshold $\theta = 0.25$ on the proportional output label change across frames, while scheduling updates based on pixel difference alone results in updating 90\% of the frames.
On the right we show output segmentations from the adaptive clockwork network as well as ground truth segments for select frames from dynamic parts of the scene (second and third frames shown) and relatively static periods (first and second frames shown).
}
\label{fig:adapt-clk-video}
\end{figure}
\setlength{\floatsep}{\the\floatsep}
\setlength{\intextsep}{\the\intextsep}
\setlength{\textfloatsep}{\the\textfloatsep}

For a closer inspection, we study one Youtube video in more depth in Figure~\ref{fig:adapt-clk-video}.
We first visualize the clock updates for our adaptive method (top left) and for a simple pixel difference baseline (bottom left), where black indicates the clock is on and the corresponding frame is fully computed.
This video has significant change in certain sections (ex: at frame $\sim 100$ there is zoom and at $\sim 350$ there is motion) with long periods of relatively little motion (ex: frames $110-130$).
While the pixel difference metric is susceptible to the changes in minor image statistics from frame to frame, resulting in very frequent updates, our method only updates during periods of semantic change and can cache deep features with minimal loss in segmentation accuracy: compare adaptive clock segmentations to ground truth (right).

\pagebreak
\section{Conclusion}
\label{sec:conclusion}
Generalized clockwork architectures encompass many kinds of temporal networks, and incorporating execution into the architecture opens up many strategies for scheduling computation.
We define a clockwork fully convolutional network for video semantic segmentation in this framework.
Motivated by the stability of deep features across sequential frames, our network persists features across time in a temporal skip architecture.
By exploring fixed and adaptive schedules, we are able to tune processing for latency, overall computation time, and recognition performance.
With adaptive, data-driven clock rates the network is scheduled online to segment dynamic and static scenes alike while maintaining accuracy.
In this way our adaptive clockwork network is a bridge between convnets and event-driven vision architectures.
The clockwork perspective on temporal networks suggests further architectural variations for spatiotemporal video processing.

\bibliographystyle{splncs}
\bibliography{2016-eccv-fast-fcn}
\end{document}


\pagestyle{headings}
\mainmatter
\def\ECCV16SubNumber{449}

\title{\emph{Supplementary Material}}
\title{Clockwork Convnets for Video Semantic Segmentation}

\titlerunning{ECCV-16 submission ID \ECCV16SubNumber}

\authorrunning{ECCV-16 submission ID \ECCV16SubNumber}

\author{Anonymous ECCV submission}
\institute{Paper ID \ECCV16SubNumber}

\maketitle
 This supplementary material accompanies submission \ECCV16SubNumber.
 In our submission we presented Clockwork FCN, an algorithm for scheduling computation in a fully convolutional network for semantic segmentation. We propose a technique for adaptively updating the clock schedule based on the input video and also explore a pipeline schedule to incorporate asynchronous updates from various layers in the convolutional network.
 In this document and the accompanying videos, we provide further qualitative examples from the Youtube-Objects dataset.
 
\section{Pipeling Scheduling Video}
We provide an additional qualitative example of the pipeline schedule in the attached video ``pipeline.mp4." 
The video simultaneously shows in order from left to right: original video, per frame oracle, pipelined (ours), and a skip frame alternative that computes the full FCN on every other frame. 
(Note this is not a fair baseline since it does not have the same reduced latency as the pipeline. However it is a useful comparison to see how the skip layers in the pipeline schedule correctly update fine spatial details for the current frame.)
The sequences of frames presented are illustrative examples of those for which our method performs better than the skip frame baseline, as measured by mean IU using the per frame oracle as ground truth. 
The sequences are drawn from a random sample of 40 shots from videos in the horse and car categories. 
The four example sequences shown demonstrate some properties of our approach. 
Fine local details are correct in real time - note the the accurate positions of the horse's legs in the first sequence, and the correct occlusion of the car in the second sequence. 
Boundaries of objects are correctly updated at each frame - note the car growing larger at every frame as it moves towards the camera in the third sequence. 
Entrance and exit of objects is detected immediately - note the segmentation tracking the car as it drives off screen in the fourth sequence.

\section{Adaptive Clockwork Results}
\begin{figure}[h!]
	\centering
	\includegraphics[width=\linewidth]{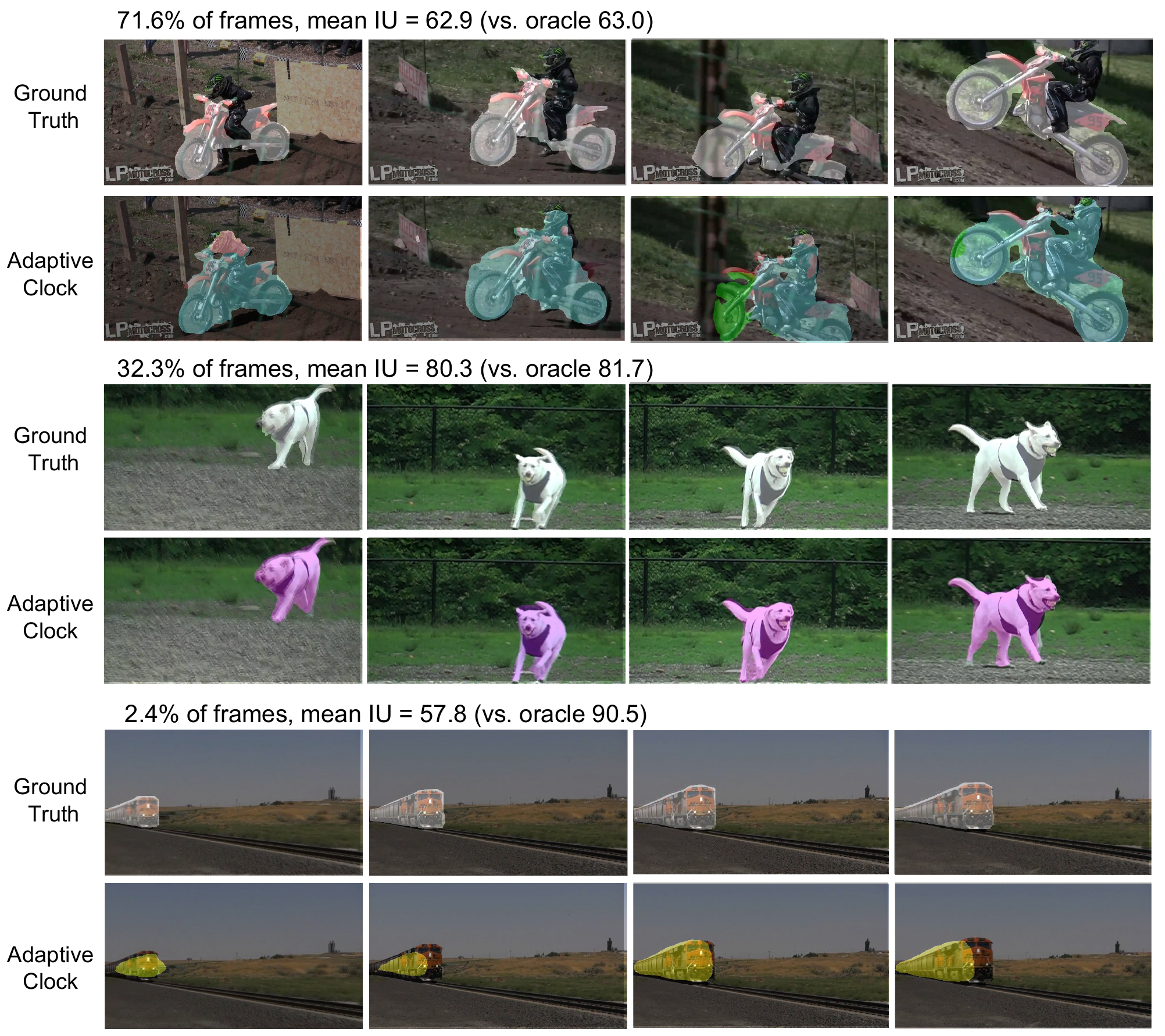}
	\caption{Illustrative examples of our adaptive clockwork method on three different videos from Youtube-Objects. We choose the threshold (on the proportional output label change across frames) $\theta = 0.25$ for the adaptive clock. For each video, the top row shows the ground truth annotations, while the bottom row shows the output of the adaptive clockwork network. Above each sequence of frames, we include the percentage of frames for which the full network was computed, as well as the mean IU score for the adaptive method compared to the oracle of the full FCN run every frame.}
	\label{fig:adapt-clk-video}
\end{figure}

We present additional adaptive clockwork results in Figure~\ref{fig:adapt-clk-video}, where we show three further examples. 
For these results, we use the threshold $\theta = 0.25$ as it represents a favorable tradeoff between accuracy and computation. 
In general, the threshold can be tuned to a particular application and dataset. 
The \emph{top row} demonstrates the performance of the adaptive clock on a video with lots of motion. The full network is updated relatively frequently (71.6\% of the time) with an almost negligible drop in accuracy (62.9\% vs 63.0\%).
The \emph{middle row} shows an example where the only object in the scene, a dog, moves quickly but generally remains in the same part of the frame: we fully update only 32.3\% of the frames causing a drop of 1.4\% in accuracy.
Finally in the \emph{bottom row}, we show an extreme example where the full network is updated on only 2.4\% of frames. 
As expected, this results in a significant drop in accuracy (57.8\% vs. the oracle 90.5\%), however, remarkably, updating only the lower layers of the network still results in updated segmentations that follow the train throughout the video while benefiting from a dramatic reduction in computation. 

Lastly, we show our clockwork FCN algorithm on a full video in ``adaptive.mp4." 
Again we use the threshold $\theta = 0.25$. This video has relatively little motion in the first half, and more motion in the second half as the camera pans. Below the video clip (original video overlaid with our segmentation) we visualize the firings of the adaptive clock as the video progresses - the blue vertical bar corresponds to the current frame while the black vertical bars correspond to the clock firing. 
During the static first half of the video, the clock fires rarely, while during the period of high motion, the adaptive clock fires more often to better track the bird as it walks.
On this video, we run the full FCN on only 8.5\% of the frames and achieve a mean IU of 69.4 (oracle mean IU is 75.3).

\setlength{\floatsep}{12pt}
\setlength{\intextsep}{12pt}
\setlength{\textfloatsep}{12pt}
\setlength{\floatsep}{\the\floatsep}
\setlength{\intextsep}{\the\intextsep}
\setlength{\textfloatsep}{\the\textfloatsep}
